\documentclass[letterpaper, 10 pt, conference]{ieeeconf}  

\IEEEoverridecommandlockouts                              

\usepackage{graphicx} 
\usepackage{times} 
\usepackage{amsmath} 
\usepackage{amssymb}  
\usepackage{bm}
\usepackage[font=small,labelfont=bf]{caption}

\usepackage{times} 
\usepackage{amsmath} 
\usepackage{amssymb}  
\usepackage[linesnumbered,ruled]{algorithm2e}
\usepackage{subfigure}
\usepackage{graphicx}
\let\labelindent\relax
\usepackage{multirow}
\usepackage[inline]{enumitem}
\usepackage[export]{adjustbox}
\usepackage[english]{babel}
\usepackage[utf8]{inputenc}

\let\oldnl\nl
\newcommand{\nonl}{\renewcommand{\nl}{\let\nl\oldnl}}
\usepackage[noend]{algpseudocode}
\usepackage{caption}
\usepackage{mathtools}

\usepackage{xcolor}
\usepackage{soul}
\usepackage{bm}
\usepackage{array}
\newcolumntype{M}[1]{>{\centering\arraybackslash}m{#1}}
\usepackage[utf8]{inputenc}
\usepackage{cite}

\title{\LARGE \bf
Multi-Robot Formation Control Using Reinforcement Learning
}

\author{Abhay Rawat and Kamalakar Karlapalem
}

\begin{document}

\maketitle
\thispagestyle{empty}
\pagestyle{empty}

\begin{abstract}

In this paper, we present a machine learning approach to move a group of robots in a formation.
We model the problem as a multi-agent reinforcement learning problem.
Our aim is to design a control policy for maintaining a desired formation among a number of agents (robots) while moving towards a desired goal. This is achieved by training our agents to track two agents of the group and maintain the formation with respect to those agents. 
We consider all agents to be homogeneous and model them as unicycle \cite{unicycle}.
In contrast to the leader-follower approach, where each agent has an independent goal, our approach aims to train the agents to be cooperative and work towards the common goal.
Our motivation to use this method is to make a fully decentralized multi-agent formation system and scalable for a number of agents.

\end{abstract}

\section{Introduction}

Multi-robot system has recently become a popular problem in robotics.
Some of the work done on multi-robot and multi agent systems are area exploration \cite{ae1} \cite{ae2}, payload transportation \cite{lcpts} \cite{yanhit}, etc. One of the classic examples of multi-robot coordination is formation control of agents, i.e. a group of agents moving while maintaining a particular shape.
Various control algorithms have been implemented to move the agents in formation which are typically based on leader-follower control architecture \cite{formation}, virtual structure \cite{vc} control architecture and behaviour modeling based control \cite{bb}.\par
Among the above mentioned architectures, leader-follower based control architecture is the most popular due to its easy implementation and robustness.
In leader-follower based approach, a leader is assigned to the formation that moves towards the desired goal using desired motion planning and tracking algorithms.
All the other agents maintain a specific distance and angle with respect to the leader robot to maintain the formation.
\par
In our work, we explore the formation control problem through a coordination based approach using reinforcement learning.
\textit{Instead of following a designated leader, the agents learn to coordinate with each other to achieve the common goal (go to goal while maintaining a given formation)}.

We also propose to make this method scalable with respect to the number of agents in formation and also the shape of the formation. The policies of the agents are trained to track any two (or any predefined number) of their neighbours. Once the policies are trained, this model can be used for different number of agents and formation shapes.
\par
In \cite{lfgq}\cite{SPRINGER667}, the authors use reinforcement learning for formation control in conjunction with leader-follower approach. The agent (robot) would learn how to maintain a specified distance from a designated leader.
\begin{figure}[h]
    \vspace{0.7em}
    \includegraphics[scale=0.4, center]{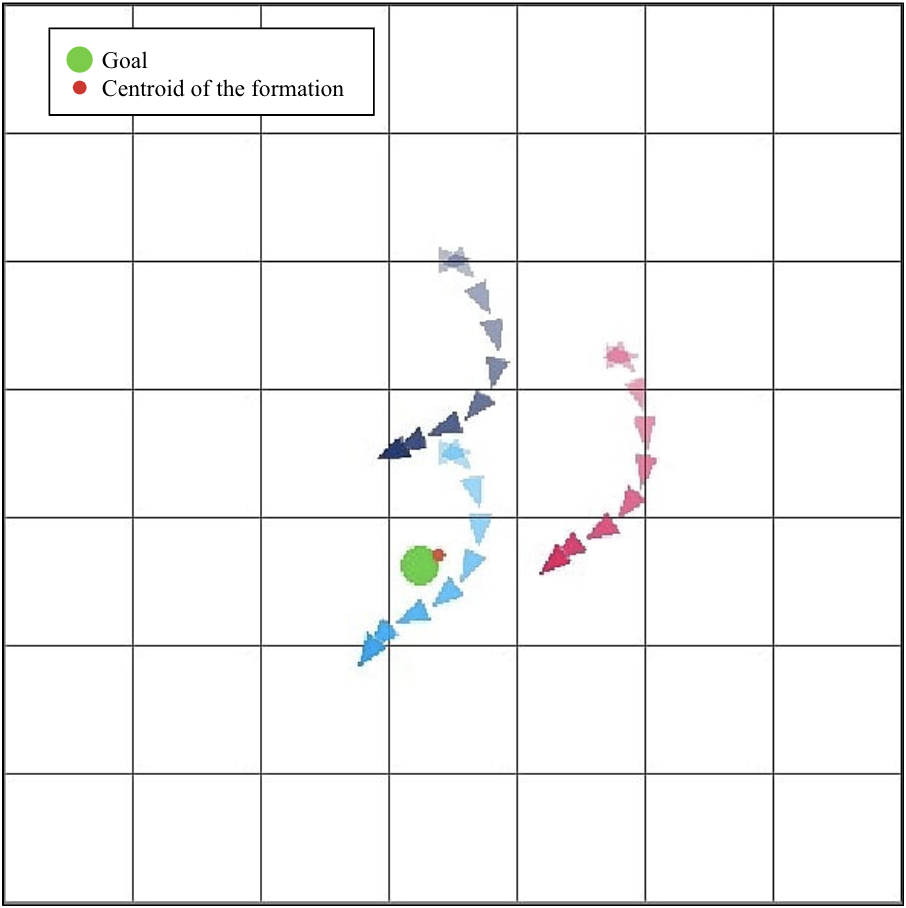}
    \caption{A multi exposure image of three agents(denoted as triangles) moving in an equilateral triangle formation towards the goal}
    \label{timelapse}
    \vspace{-1em}
\end{figure}
In \cite{johns2018intelligent}, the authors use reinforcement based control alongside a behaviour based formation controller. The actions of the agent would be the average of the control signals of both the controllers. In both of the above mentioned cases, the agents work towards individual goals to complete the task, whereas in our work we aim to develop coordination between agents to achieve that common goal.
Unlike \cite{turpin2012decentralized} and \cite{cheng2014decentralized}, we do not consider communication amongst the agents and investigate the effectiveness of our method without it.


\section{Problem Formulation}

We formulate the task of distributed formation control as a discounted stochastic game $G$, described by the tuple $G = \langle S, U, A, P, R, Z, O, \gamma \rangle$.
The true state of the environment is denoted by $s \in S$, where $S$ denotes the state space of the environment.
At each time-step every agent $a \in A$ takes an action $u_{a} \in U$ resulting in the transition of the environment to some other state $s^{\prime}$.
The transition is stochastic, and the probability of the next state $s^{\prime}$ conditioned on the actions $u$ and the state $s$ is given by $P(s^{\prime}|s, \bar{u})$: $S \times \bm{U} \times S \rightarrow [0, 1]$, where $\bar{u} \in \bm{U}$ is the joint action of all the agents and $\bm{U} = U^{|A|}$ is the joint action space. Due to the cooperative nature of the task all the agents share a common reward function $R(s, \bar{u}): S \times \bm{U} \rightarrow \mathbb{R}$ and $\gamma \in [0, 1]$ is the discount factor. At any given time-step, an agent $a$ observes the partial state of the environment $o \in O$ according to the observation function $Z(s, a): S \times A \rightarrow O$.\par
Each agent interacts with the environment according to it own policy. 
Each agent has its history of interactions with the environment which it uses to train its policy.
Following the work of the authors from \cite{COMA} \cite{MADDPG}, we use the paradigm of centralized training with decentralized execution.
We use the actor-critic architecture with a central critic that aims to approximate the state-action value function.
The central critic would represent a state action value function which maps the true state of the environment and the joint actions, to the expected sum of rewards.
We use a deep neural network $Q$ parameterized by $\theta_{q}$ to approximate this function. In particular, the critic function aims to minimize the following loss:
\begin{align}
    J_{Q}(s, \bar{u}) =  \left\lVert \hat{y} - Q(s, \bar{u} | \theta_{q}) \right\rVert_{2}
    \label{qloss}
\end{align}
where $\hat{y} = R(s, \bar{u}) + \gamma \;\underset{\bar{u}^{\prime}}{max} \, \bar{Q}(s^{\prime}, \bar{u}^{\prime})$, $\bar{u}^\prime$ is the next joint action taken by the agents according to their current policy, and $\bar{Q}$ is the target network (parameterized by $\bar{\theta}_q$) as used by \cite{ddqn}.
The parameters of the target critic network are updated by taking a weighted average of the parameters of both the critic networks (polyak averaging \cite{polyak}).
The weights of the target critic network are updated as
\begin{align}
    \bar{\theta_{q}} = \tau\theta_q + (1-\tau)\bar{\theta_q}
\end{align}
where $\tau \ll 1$.
This update (soft update) has been found to stabilize the learning process \cite{DDPG} by reducing the variance in the target $Q$ values used to estimate the loss in Eq. \ref{qloss}.\par
\begin{figure}[h]
    \includegraphics[scale=.25, center]{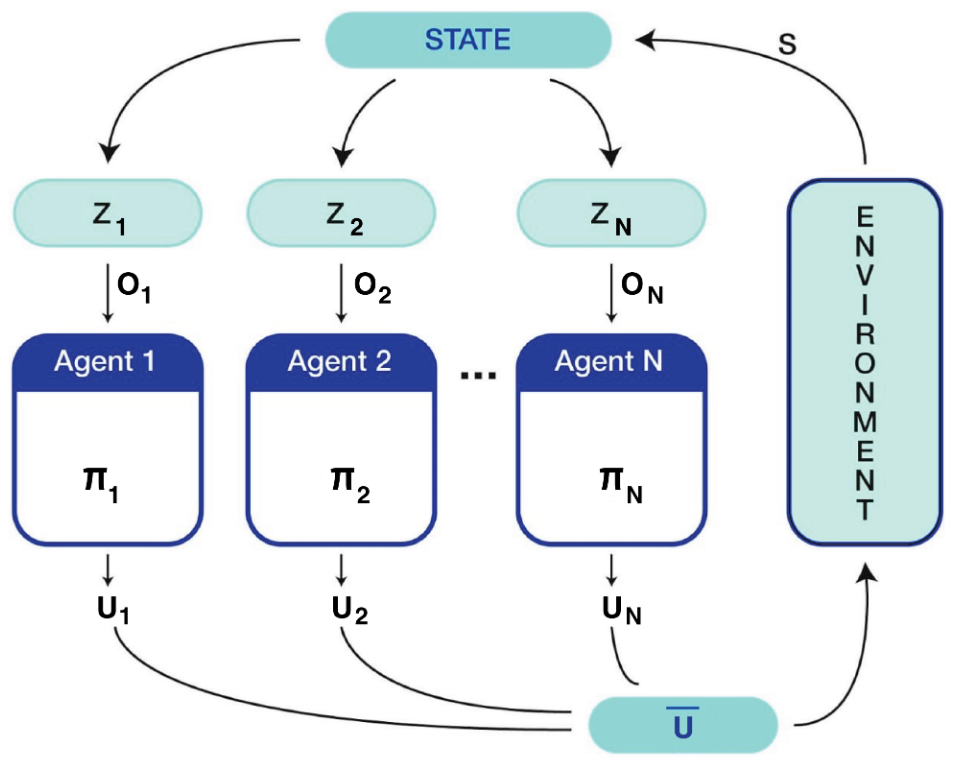}
    \caption{Interaction of agents with the environment. The state of the environment is observed through the observation function $Z_{i}$ of the agent $i$. 
    The agent acts according to its policy $\pi_{i}$. The actions of all the agents $\bar{u}$, transitions the environment to some new state.}
    \label{model}
\end{figure}
We use policy gradient method \cite{dpg}\cite{DDPG} to train the actor network $\pi$, parameterized by $\theta_{\pi}$, by adjusting its parameters ($\theta_{\pi}$) to maximize the expected reward $J_{\pi}$, where
\begin{equation*}
    J_{\pi}(\cdot, \theta_{\pi}) = \mathbb{E}_{s \sim p^{\pi}, a \sim \pi}[R(s, \bar{u})]
\end{equation*}
Every agent's policy is trained by updating its parameters in the direction of $\nabla_{\theta_i} J_{\pi}$ (gradient of the loss $J_{\pi}$ with respect to the parameters of $\pi_i$). 
Policy gradient methods are known to result in high variance estimates of the gradient. In multi-agent setting this is quite an issue as the agents gradients are also dependent on the action of other agents \cite{MADDPG}. To mitigate this problem to some extent we stick to the deterministic flavour of the algorithm proposed in \cite{DDPG}.\par
For distributed formation control, we leverage the notion of minimal rigidity and persistent graphs \cite{hendrickx2005rigidity} to define our system.
We denote each agent as a vertex $v_{i} \in V$ in a graph $\mathbb{G} = (V, E)$. There is a directed edge $e_{i, j} = (v_{i}, v_{j}) \in E$ between two agents $v_{i}, v_{j}$ if and only if agent $v_{i}$ observes agent $v_{j}$.
Following \cite{hendrickx2005rigidity}, we use the fact that if each of the agent in the formation observes at least two other agents then we have a minimally rigid graph.
We can use these trained agents for any shape of formation by providing them the inter-agent distances.
In addition to observing other agents, all the agents continuously track the goal position.
The agents are rewarded if they are able to maintain the specific distance with the agents that they can observe.
A reward of greater magnitude is given if the agents are able to reach the goal while maintaining said formation.
To define the reward function, we mathematically introduce the following conditions for judging actions at the given states.

\begin{itemize}
    \item [{[C1]}] \textbf{Formation condition}: If all the agents maintain their specified edge lengths within an error of $\epsilon_{form}$
    \begin{equation}
        \left\lvert d_{i, j}(s) - \overline{d}_{i, j} \right\rvert \leq \epsilon_{form} \quad  \forall i, j \in E
        \label{form_condition}
    \end{equation}
    \item [{[C2]}]\textbf{Collision condition}: This condition is met if the distance between any two agents is less than some threshold (i.e., collision). $\epsilon_{col}$.
    \begin{equation}
        \left\lvert d_{i, j}(s) \right\rvert \leq \epsilon_{coll} \quad  \forall i, j \in E
        \label{collision_condition}
    \end{equation}
    \item [{[C3]}]\textbf{Success condition}: If formation condition is satisfied and
    
    \begin{equation}
        \left\lvert c_{V} - g\right\rvert \leq \epsilon_{goal}
        \label{term_condition}
    \end{equation}
\end{itemize}
where $d_{i, j}(s)$ represents the distance between agents $i$ and $j$ at state $s$, $\overline{d}_{i, j}$ denotes the desired distance between agents $i$ and $j$, $c_{V}$ denotes the centroid of the formation made by the agents in $V$, $g$ is the desired goal point of the formation, $\epsilon_{form}$ and $\epsilon_{goal}$ are the tolerances in error. The reward function is defined as:
\begin{align}
    R(s, a, s^{\prime}) = 
    \begin{cases}
        r_{edge}, & \text{if only [C1] is true}\\
        r_{collision}, & \text{if [C2] is true}\\
        r_{goal}, & \text{if [C3] is true}\\
        r_{penalty}, & \text{otherwise}
    \end{cases}
\end{align}
\par
To encourage the agents to complete the task as soon as possible, we give a small negative penalty $r_{penalty}$. We also ensure that $r_{edge} \ll r_{goal}$ so that the agents do not get stuck in a local minima (where they remain in formation while totally ignoring the high level goal). The values for these parameters that were used in our experiments are described in Table \ref{param}.

\begin{table}[h]
\centering
\captionsetup{ width= 65mm}
\captionof{table}{List of Parameters.}

\renewcommand{\arraystretch}{1.3}
\begin{tabular}{ |M{4cm}||M{2cm}|  }
 \hline
 Parameter Name & Value\\
 \hline
 H    &15\\
$\epsilon_{form}$ &0.10 m\\
$\epsilon_{goal}$ &0.15 m\\
$\epsilon_{coll}$ &0.20 m\\
$r_{edge}$ & 0.1\\
$r_{collision}$ & -100\\
$r_{goal}$ & 50\\
$r_{penalty}$ & -0.5\\
$\tau$ &5e-3\\
 \hline
\end{tabular}
\label{param}
\vspace{-1em}
\end{table}
\section{Methodology}
The control inputs of each agent is its linear and angular velocity ($v$ and $\omega$).
Each agent $i$ observes up to two other agents along with the target location.
The observations are made in the local reference frames of the observing agents.
Each agent has its own independent policy $\pi$ which maps the observation space to the action space. We use a neural network to define this function with parameters $\theta_{\pi}$.
To mitigate the problem of partial observability we provide the agents a sequence of observations of length $H$.
This helps the agents to learn the behaviour of other agents.
In particular, the actor function (policy) can be described as:
\begin{equation*}
\pi(o_{t-H-1}, \ldots, o_{t}): O^{H} \rightarrow U
\end{equation*}

We define a central critic \cite{COMA} that maps the true state of the system and the joint actions taken by the agents at that state to the expected future return (state-action value).
The true state of the system is defined as the pose information of all the agents with respect to the centroid of the current formation i.e. $[x^{c}_{i}, y^{c}_{i}, \theta^{c}_{i}] \; \forall i \in V$ augmented with the position vector of the goal state with respect to the centroid.
The critic function is described as:
\begin{equation*}
Q(s_{t-H-1}, \ldots, s_{t}, \bar{u}_{t-H-1}, \ldots, \bar{u}_{t}): S^{H} \times U^{H} \rightarrow U
\end{equation*}

\begin{figure}[h]
    \vspace{-4em}
    \includegraphics[scale=0.07, center]{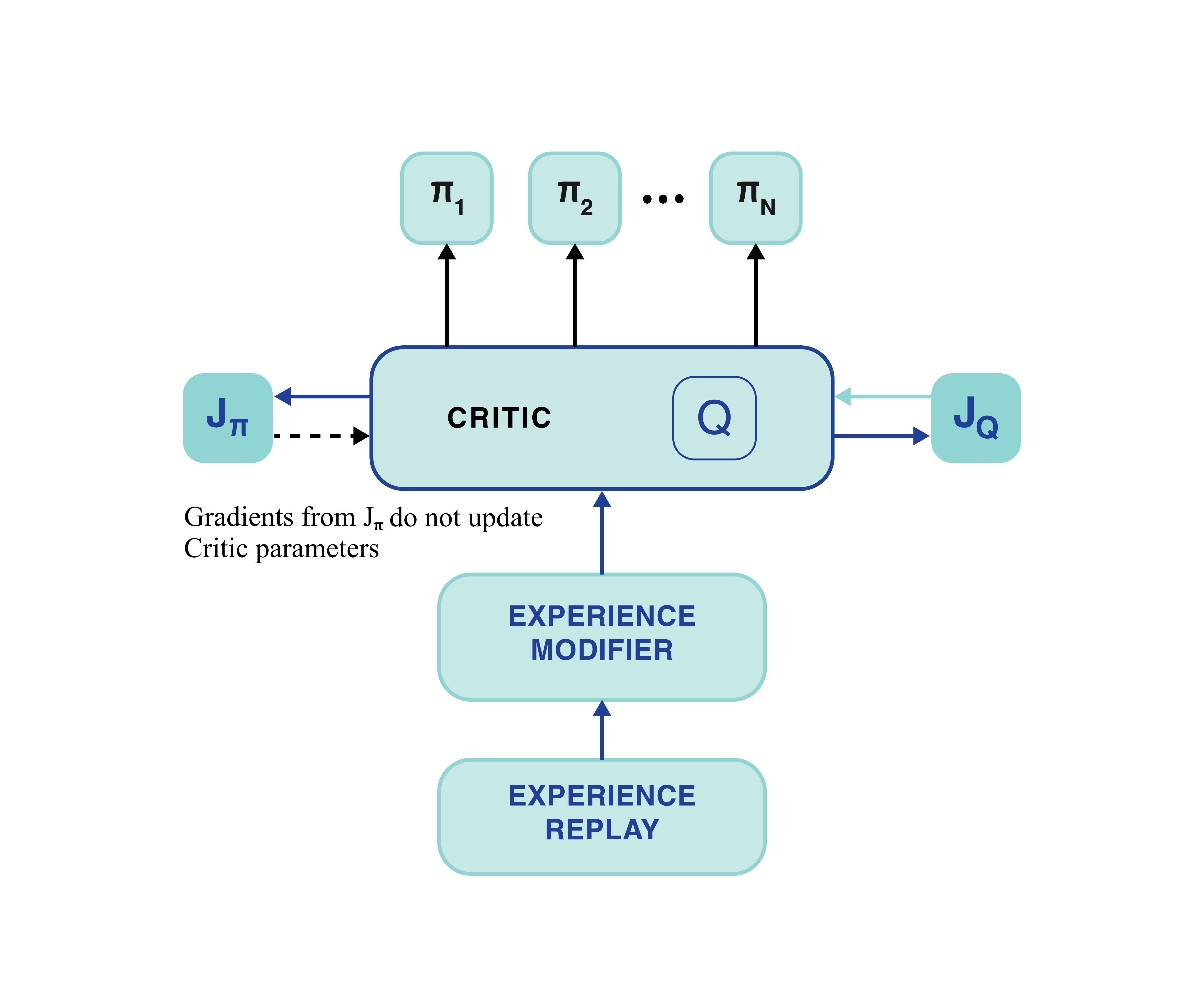}
    \vspace{-3em}
    \caption{The training involves sampling a batch of experiences, then modifying some of the experiences for positive reinforcement.
    }
    \label{architecture}
\end{figure}

We build a centralized experience replay containing tuples of the past experiences  $\langle s, r, s^{\prime}, u_{1}, \ldots, u_{|A|}, o_{1}, \ldots, o_{|A|}, o^{\prime}_{1}, \ldots, o^{\prime}_{|A|} \rangle$, where $s^{\prime}$ and $o^{\prime}$ are the next state of the environment and the observation made by the agent $i$ for the corresponding state $s^{\prime}$ respectively.
At each training iteration we randomly sample a sequence of $H$ concurrent experiences from this memory.
We use a similar approach as proposed by the authors of \cite{CERT}, to sample concurrent experiences.
\par
To speed up learning, we modify the batch of experiences using a similar method as in \cite{HER}.
This is shown as experience modifier in Fig. \ref{architecture}.
The authors in \cite{HER} change the goal state  to the actual state achieved by the agent.
This leads in more positive reinforcement and thus speeds up the learning process considerably.
In this case, the goal is observed with respect to different observation frames and the input to the networks is a sequence containing those goals at different time-steps.
In order to make the changes consistent throughout the sequence, we keep track of the transformations from previous pose to the current pose, for every agent.
For a sequence of $H$ observations $o_{1}, ...o_{H}$, let the goal achieved be $g_{t}$, and the transformation matrix between the consecutive poses of an agent be $K_{t}^{t+1}$, then:
\begin{align}
g_{t} = K_{t}^{t+1} \times g_{t+1} \qquad t \in 1, \ldots, H-1
\end{align}
where $g_{H}$ is the centroid of the formation at time-step $H$.
\par

\section{Results}
The simulations are performed using three agents.
The agents are given a desired formation to maintain and a desired goal. During execution, only the actor network is used, which can be easily executed on low cost mobile computers such as Raspberry Pi.
\begin{figure}[t!]
    \vspace{-1em}
    \includegraphics[scale=0.6, center]{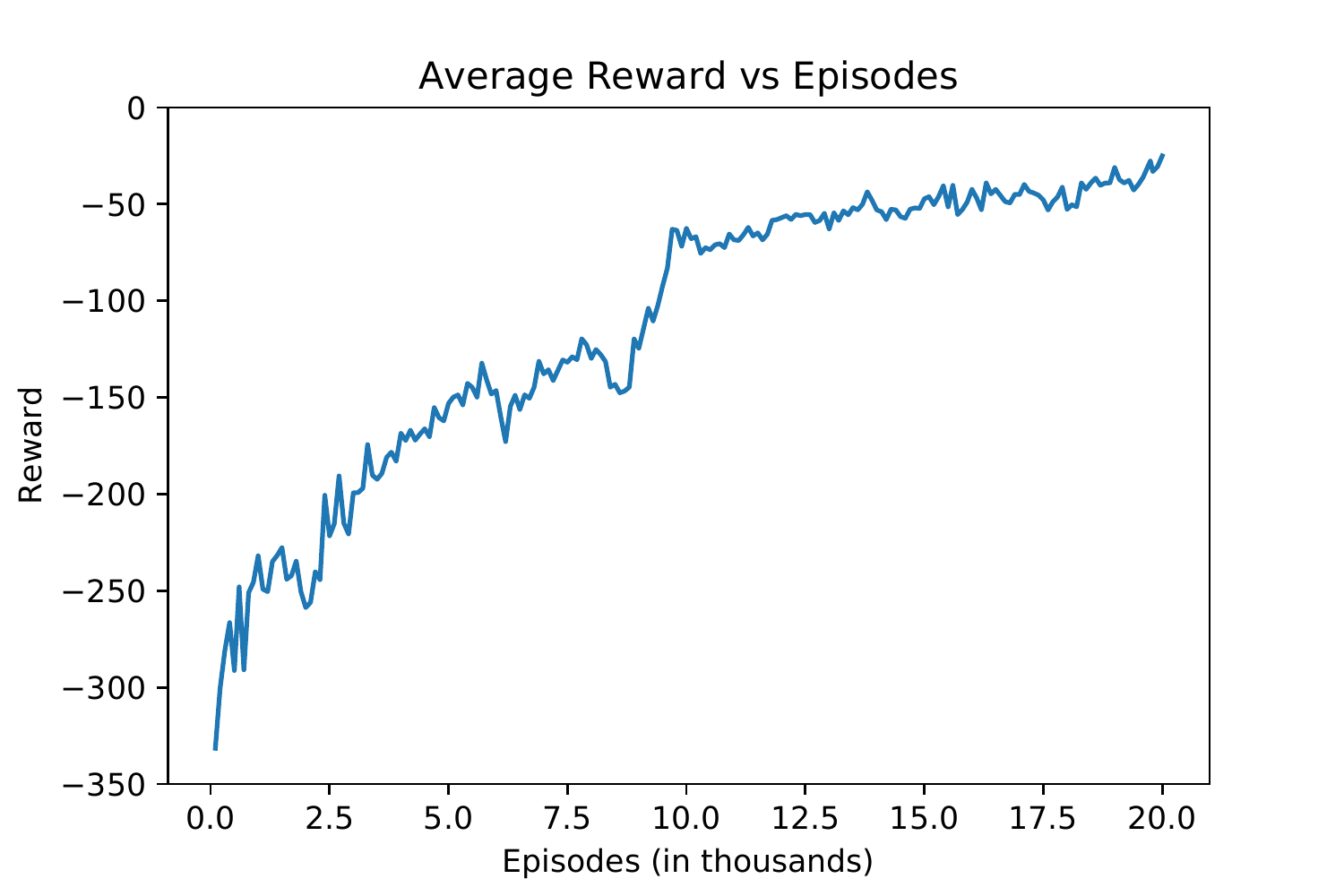}
    \caption{Average reward per episode vs. number of training episodes}
    \label{learning}
    \vspace{-1em}
\end{figure}

\par
Fig. \ref{learning} shows the learning curve for training.
The flow of the training is described in Fig. \ref{architecture}.
We trained the agents for $20000$ episodes, where each episode had a maximum of 120 time-steps.
An episode could terminate early if there is any collision amongst the agents, or task was completed within $120$ time-steps.
The environment is a custom environment made with the help of OpenAI gym \cite{gym}.
The arena is a square of side length $10\text{ m}$.
During training, the goal state, the centroid of the formation and the formation to be maintained were generated at random.
To facilitate the training process, The initial states of individual agents were computed from the centroid and the target formation to make sure that the agents were in formation in the initial state.
\begin{figure}[h]
    \includegraphics[scale=0.35, center]{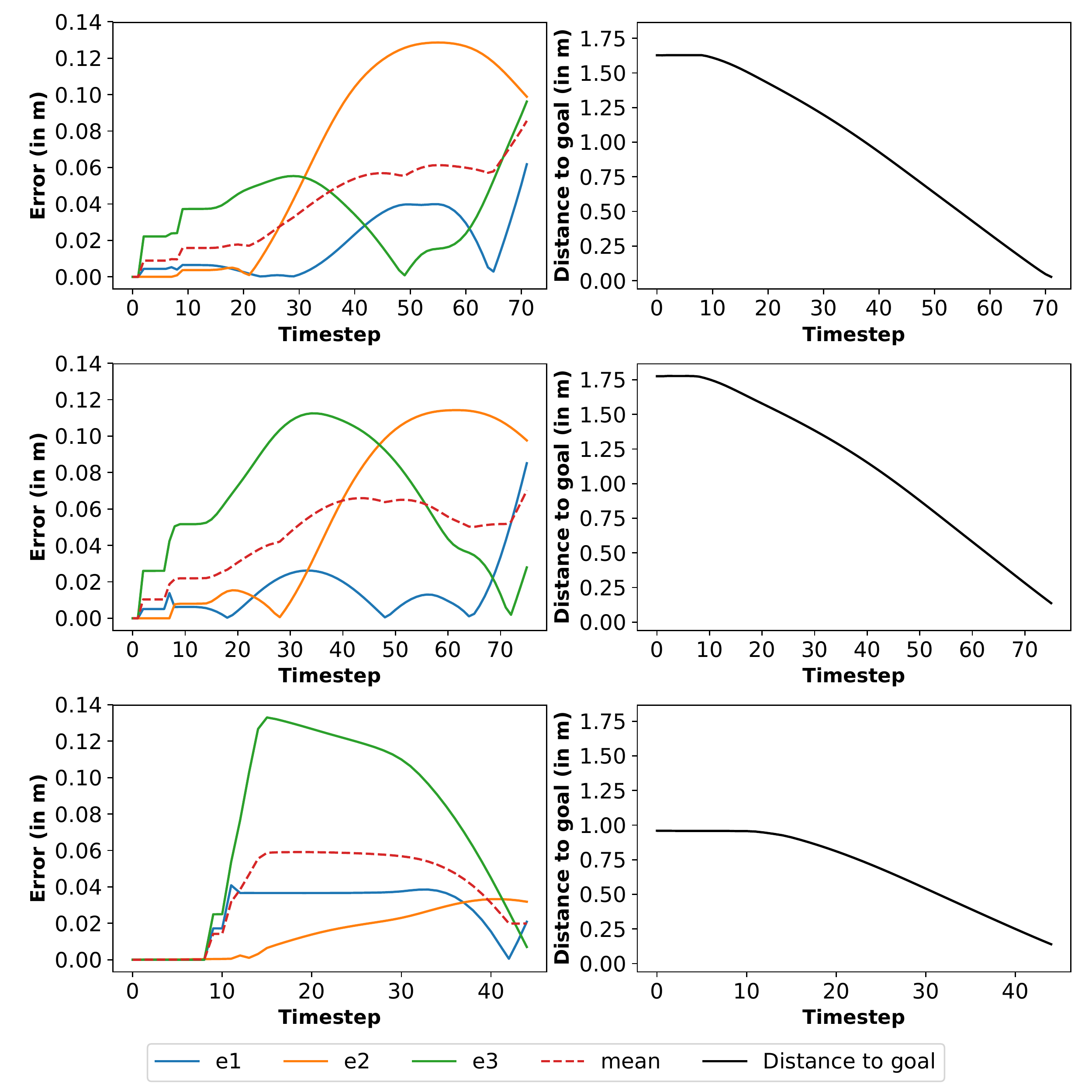}
    \caption{The plots on the left show the error in each edge of the formation.
    The plots on the right show the distance between the centroid of the formation and the goal. The edges between agents are denoted by $e1$, $e2$ and $e3$. The dotted red line denotes the mean error of all the edges.}
    \label{error}
\end{figure}
\par
Fig. \ref{error} shows the error in each edge of the formation
alongside the distance of the centroid from the goal with respect to time for the corresponding trial.
\par
\section{Conclusion}
In contrast to existing formation control algorithms, we aim to build a scalable and modular control mechanism for formations through multi-agent reinforcement learning.
\par
Multi-agent systems suffer from non-stationarity of the environment with respect to the agents. This can cause fluctuations in the learning process as well as during execution.
In a task like formation control, this can result in cascading effect and the agents might not be able to recover from a poor choice of action.
This effect can be seen in the first two rows of Fig. \ref{error}, where the error in the edges fluctuates throughout the episode.
In future work we aim to improve the accuracy of the formation and to address the problem of non-stationarity which hinders the performance when more agents are added.
To mitigate this issue we can limit the change in policy at every iteration for each agent.
Such method has already been used by \cite{PPO} \cite{DBLP:journals/corr/SchulmanLMJA15} to prevent the policy to change drastically.
By doing this we hope that the agents would be able to learn about the policies of other agents and take actions accordingly.

\bibliographystyle{ieeetr}
\bibliography{root}
\end{document}